\documentclass[sigconf,nonacm]{acmart}
\AtBeginDocument{%
  }

\usepackage{graphicx}
\usepackage[skip=2pt]{caption}
\usepackage{booktabs}
\usepackage{multirow}
\usepackage{xcolor}
\usepackage{tcolorbox}
\usepackage{lipsum}
\usepackage{amsmath}

\usepackage{amssymb}
\usepackage{tikz}
\usepackage{subcaption}
\usepackage[hang,flushmargin]{footmisc}
\usepackage{float}
\usepackage[ruled,vlined,linesnumbered]{algorithm2e}

\begin{document}

\title{Reference-Aligned Retrieval-Augmented Question Answering over Heterogeneous Proprietary Documents}

\author{Nayoung Choi}
\affiliation{%
  \department{Department of Computer Science}
  \institution{Emory University}
  \city{Atlanta}
  \state{Georgia}
  \country{USA}
}
\email{nayoung.choi@emory.edu}

\author{Grace Byun}
\affiliation{%
  \department{Department of Computer Science}
  \institution{Emory University}
  \city{Atlanta}
  \state{Georgia}
  \country{USA}
}
\email{grace.byun@emory.edu}

\author{Andrew Chung}
\affiliation{%
  \department{Department of Computer Science}
  \institution{Emory University}
  \city{Atlanta}
  \state{Georgia}
  \country{USA}
}
\email{andrew.chung@emory.edu}

\author{Ellie S. Paek}
\affiliation{%
  \department{Department of Computer Science}
  \institution{Emory University}
  \city{Atlanta}
  \state{Georgia}
  \country{USA}
}
\email{ellie.paek@emory.edu}

\author{Shinsun Lee}
\affiliation{%
  \institution{Hyundai Motors}
  \city{Seoul}
  \country{South Korea}
}
\email{sslee333@hyundai.com}

\author{Jinho D. Choi}
\affiliation{%
  \department{Department of Computer Science}
  \institution{Emory University}
  \city{Atlanta}
  \state{Georgia}
  \country{USA}
}
\email{jinho.choi@emory.edu}

\renewcommand{\shortauthors}{Choi et al.}

\begin{abstract}
Proprietary corporate documents contain rich domain-specific knowledge, but their overwhelming volume and disorganized structure make it difficult even for employees to access the right information when needed. For example, in the automotive industry, vehicle crash-collision tests—each costing hundreds of thousands of dollars—produce highly detailed documentation. However, retrieving relevant content during decision-making remains time-consuming due to the scale and complexity of the material. While Retrieval-Augmented Generation (RAG)-based Question Answering (QA) systems offer a promising solution, building an internal RAG-QA system poses several challenges: (1) handling heterogeneous multi-modal data sources, (2) preserving data confidentiality, and (3) enabling traceability between each piece of information in the generated answer and its original source document. To address these, we propose a RAG-QA framework for internal enterprise use, consisting of: (1) a data pipeline that converts raw multi-modal documents into a structured corpus and QA pairs, (2) a fully on-premise, privacy-preserving architecture, and (3) a lightweight reference matcher that links answer segments to supporting content. Applied to the automotive domain, our system improves factual correctness (+1.79, +1.94), informativeness (+1.33, +1.16), and helpfulness (+1.08, +1.67) over a non-RAG baseline, based on 1–5 scale ratings from both human and LLM judge. The system was deployed internally for pilot testing and received positive feedback from employees.

\end{abstract}




\maketitle

\section{Introduction}


\begin{figure*}[t!]
    \centering
    \includegraphics[width=1.0\linewidth]{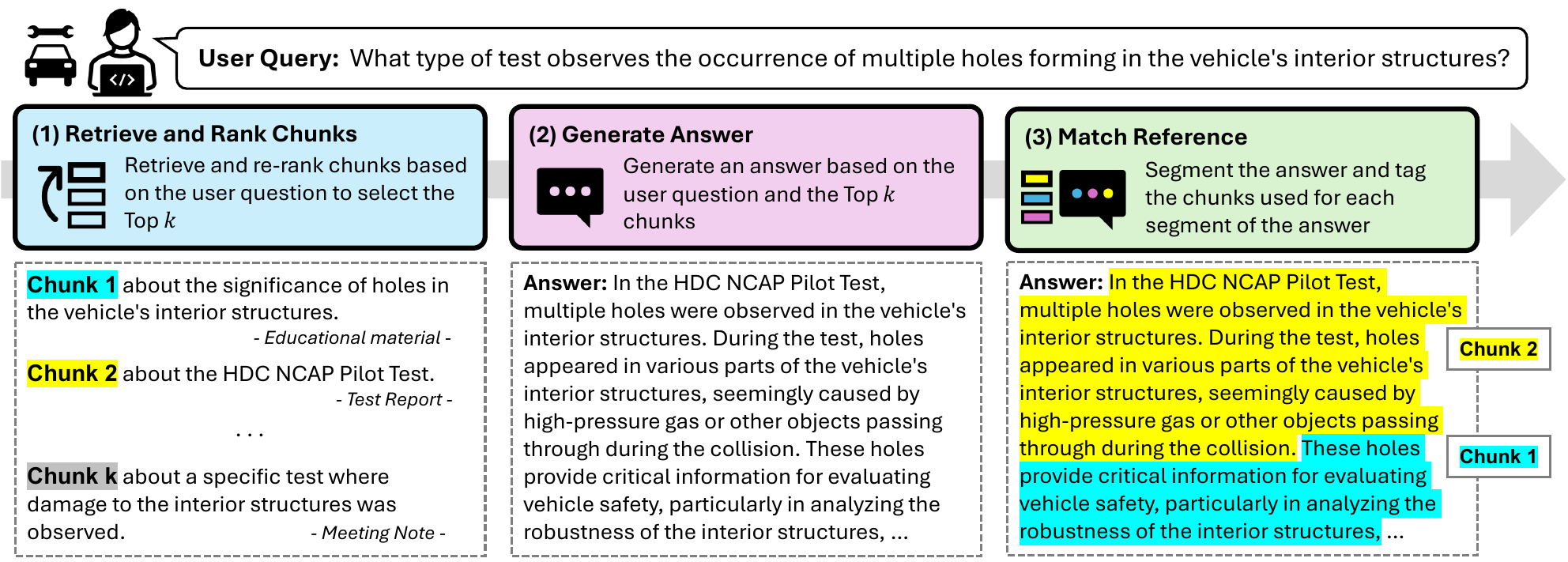}
    \vspace{-0.4cm}
    \caption{Given a user query, our RAG system retrieves relevant chunks, generates a response, and aligns each segment with its source for traceability. The example was originally in Korean and translated into English.}
    \label{fig:RAG}
    \vspace{-0.3cm}
\end{figure*}

Retrieval-Augmented Generation (RAG) has shown potential in reducing hallucinations and providing up-to-date knowledge in Large Language Models (LLMs). This success has grown interest in domain expert RAG-Question Answering (QA) systems to meet specialized knowledge needs. While previous studies \cite{han-etal-2024-rag} \cite{siriwardhana-etal-2023-improving} \cite{mao-etal-2024-rag} have proposed general methods for adapting RAG models to domain knowledge bases—such as syntactic QA pair generation or model fine-tuning—they face several challenges for internal use in enterprise settings. Internal corporate documents are often presented in heterogeneous formats and exhibit an unstructured nature, making their direct integration into RAG system development challenging \cite{hong-etal-2024-intelligent}. Moreover, general-purpose LLMs often lack sufficient knowledge of specialized corporate domains \cite{zhao2024optimizing}, making fine-tuning essential. However, this process is further complicated by the scarcity of structured training data. In addition, data privacy and security concerns restrict the full utilization of API-based LLMs \cite{achiam2023gpt} \cite{claude2024} within RAG systems, necessitating the use of open-source alternatives \cite{touvron2023llama} \cite{qwen}. These limitations highlight the need for a RAG-QA pipeline tailored for internal enterprise use, one that can process heterogeneous documents while ensuring data privacy.

The automotive industry is one of the domains where an internal RAG-QA pipeline is particularly needed, due to both the scale and sensitivity of engineering data. For example, a single vehicle crash‑collision test~\footnote{A vehicle crash-collision test involves driving a car into a fixed barrier to measure occupant and pedestrian safety; each test costs hundreds of thousands of dollars.} produces hundreds of pages of confidential artifacts, including design rationales, sensor logs, finite‑element analyses. Engineers review this documentation repeatedly when diagnosing failures, refining component designs, or compiling regulatory submissions, yet the material is fragmented across file shares and internal platforms, making retrieval slow. An internal RAG-QA system enables users to retrieve necessary information without manually navigating large volumes of documents, and to query LLMs without concerns about data confidentiality—ultimately reducing the time required for decision-making.

In this work, we address three key challenges of building a RAG-QA system for internal use and present an on-premise framework, demonstrating its application in a leading automotive company based in Korea. First, to handle heterogeneous and multi-modal documents, we present a data processing pipeline (Section~\ref{sec:data_processing}) that converts raw inputs, such as meeting notes, test reports, and educational materials, into a structured corpus and the corresponding QA pairs. Second, to preserve data confidentiality, we propose an on-premise RAG system (Section~\ref{sec:approach_components}) composed of domain-adapted retrieval, re-ranking, and generation components, none of which rely on external servers. Third, to support verifiable decision-making, we introduce a lightweight reference matching algorithm (Section~\ref{sec:reference_matching_algorithm}) that links each answer segment to its supporting document, allowing users to trace the original sources and validate the information. The workflow of our RAG system is shown in Figure~\ref{fig:RAG}, and our key contributions are summarized as follows:
\vspace{-0.1em}
\begin{itemize} 
\item We present a data processing pipeline that converts heterogeneous multi-modal documents into Markdown text chunks and generates Q\&A pairs to train the RAG system.
\item We develop a framework that domain-adapts the retrieval, re-ranking, and generation components of the RAG system and deploys them in an on-premise system.
\item We introduce a reference matching algorithm that uses dynamic programming to align answer segments with retrieved documents, enabling traceability to original sources.
\end{itemize}
\vspace{-0.3em}

\section{Related Work}
\textbf{Domain-specific RAG} $\;$ RAG-Studio \cite{mao-etal-2024-rag} employs synthetic data generation for in-domain adaptation, reducing reliance on costly human-labeled datasets. However, it assumes access to well-structured data, limiting its applicability to real-world cases with unstructured raw inputs. To address such challenges, \citet{hong-etal-2024-intelligent} handle diverse formats (e.g., DOC, HWP) by converting documents to HTML while preserving structural cues, and \citet{guan-etal-2024-bsharedrag} mitigate the issue of short, noisy e-commerce data by constructing a richer dataset from raw crawled corpora. Building on these efforts, recent multimodal and enterprise-focused RAG systems—such as VisDoMRAG, which integrates visual and textual retrieval for visually rich documents~\cite{Suri2025VisDoMRAG}, and eSapiens, which combines Text-to-SQL with RAG in enterprise settings~\cite{Shi2025eSapiens}—further underscore the need for robust RAG methods capable of handling heterogeneous data.

\vspace{1em}
\noindent \textbf{RAG Evaluation} $\;$  The evaluation of RAG systems \cite{yu2024evaluationretrievalaugmentedgenerationsurvey} has relied on text similarity-based metrics such as BLEU \cite{papineni2002bleu}, ROUGE \cite{lin2004rouge}, and EM \cite{rajpurkar2016squad}. These metrics provide a baseline but overlook valid answer diversity and qualitative aspects like factual consistency and relevance. Recent advancements are made to utilize LLM-as-a-judge \cite{zheng2023judgingllmasajudgemtbenchchatbot} to evaluate qualitative dimensions. \citet{han-etal-2024-rag} propose a pairwise preference comparison framework considering helpfulness and truthfulness, while \citet{saad-falcon-etal-2024-ares} assess context relevance, answer faithfulness, and query relevance, addressing hallucination issues. However, a standardized framework for RAG evaluation remains a challenge, and the design of evaluation metrics should be tailored to the specific application context.

\vspace{0.1cm}

\section{Approach}
\subsection{Data Generation \& Processing}
\label{sec:data_processing}
\begin{figure}[htp!]
    \centering
    \includegraphics[width=1.0\linewidth]{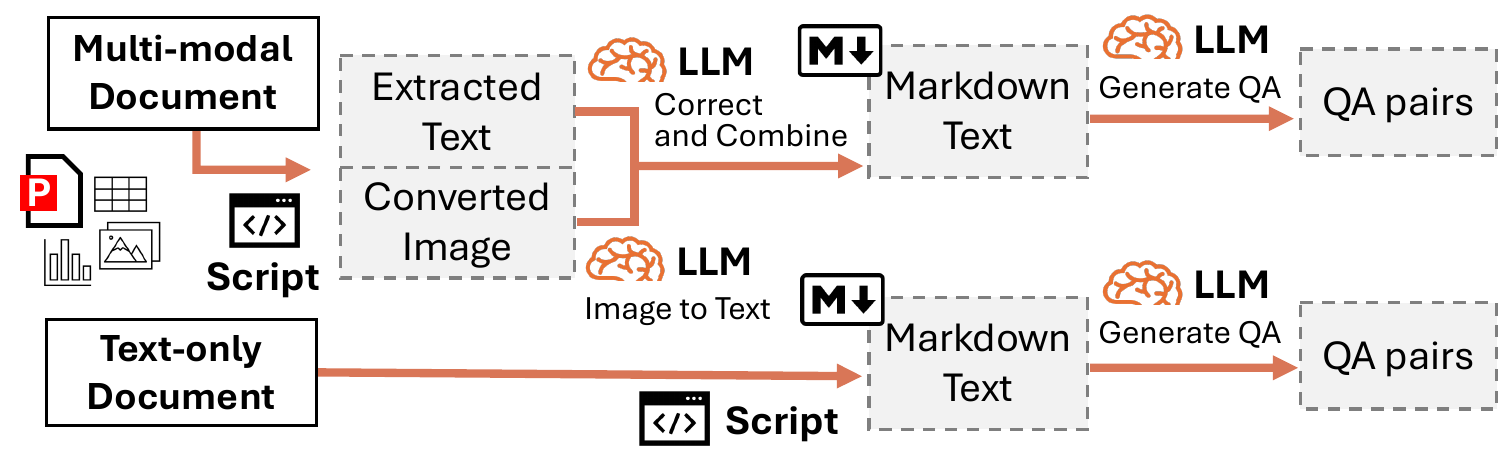}
    \caption{Overview of the data generation pipeline. Multi-modal slides are processed through image and text extraction followed by LLM-based correction, while textbook pages are directly parsed into text. All content is then converted into Markdown chunks and used to generate Q\&A pairs.}
    \label{fig:data_pipeline}
\end{figure}

We use three different internal data sources from a Korean automotive company: (1) meeting notes, (2) test reports on vehicle crash-collision tests—both in presentation slide format—and (3) the educational text material \textit{Crash Safety of Passenger Vehicles} \cite{mizuno2016crash}. The first two sources contain a mix of multi-modal elements, including images, tables, and graphs. Figure \ref{fig:data_pipeline} illustrates the process of extracting, analyzing and converting slides and textbook PDFs into structured Markdown text and Q\&A sets, leveraging Python scripts and the Claude \cite{claude2024} Bedrock API~\footnote{Amazon Bedrock service ensures all prompts and responses remained within a secure enterprise-grade environment, aligning with our data privacy requirements.}. The converted Markdown text is segmented into chunks—each slide is treated as one chunk for the slide sources, and each chapter as one chunk for the textbook. Table~\ref{tab:dataset_raw} presents the data statistics by source, while Table~\ref{tab:dataset_split} shows the dataset split into training, validation, and test sets. Token length statistics for the training data are provided in Table~\ref{tab:data_token_stat}.

\begin{table}[h!]
\centering
\begin{tabular}{llrrr}
\toprule
\multicolumn{1}{c|}{\textbf{Type}}        & \multicolumn{1}{c|}{\textbf{Source}} & \multicolumn{1}{c}{\textbf{File}}   & \multicolumn{1}{c}{\textbf{Slide}} & \multicolumn{1}{l}{\textbf{Q\&A Pairs}} \\ \midrule
\multicolumn{1}{l|}{\multirow{2}{*}{PPT}} & 
\multicolumn{1}{l|}{Test Report}     & 1,463                               & 4,662                              & 59,402                                  \\
\multicolumn{1}{l|}{}                     & \multicolumn{1}{l|}{Meeting Note}  & 249                                 & 882                                & 7,696                                   \\ \midrule
\multicolumn{1}{c|}{}                     & \multicolumn{1}{c|}{}                & \multicolumn{1}{c}{\textbf{Page}} & \multicolumn{1}{c}{\textbf{Chapter}}  & \multicolumn{1}{c}{\textbf{Q\&A Pairs}} \\ \midrule
\multicolumn{1}{l|}{PDF}                  & \multicolumn{1}{l|}{Textbook}        & 404                                   & 81                                 & 1,505                                   \\ \midrule
\multicolumn{2}{r}{\textbf{Total}}                                               & \multicolumn{1}{l}{}                & 5,625                              & 68,603                                  \\ \bottomrule
\end{tabular}%
\caption{Raw data statistics by source, along with the corresponding number of generated Q\&A pairs.}
\label{tab:dataset_raw}
\end{table}

\begin{table}[h!]
\begin{tabular}{l|l|rrr|r}
\toprule
\multicolumn{1}{c|}{\textbf{Source}} & \multicolumn{1}{c|}{\textbf{Data}} & \multicolumn{1}{c}{\textbf{Train}} & \multicolumn{1}{c}{\textbf{Val}} & \multicolumn{1}{c|}{\textbf{Test}} & \multicolumn{1}{l}{\textbf{Total}} \\ \midrule
\multirow{2}{*}{Test Report}         & Chunk                              & 3,729                              & 466                              & 467                                & 4,662                              \\
                                     & Q\&A Pair                          & 47,660                             & 5,823                            & 5,919                              & 59,402                             \\ \midrule
\multirow{2}{*}{Meeting Note}      & Chunk                              & 705                                & 88                               & 89                                 & 882                                \\
                                     & Q\&A Pair                          & 6,144                              & 752                              & 800                                & 7,696                              \\ \midrule
\multirow{2}{*}{Textbook}            & Chunk                              & 64                                 & 8                                & 9                                  & 81                                 \\
                                     & Q\&A Pair                          & 1,182                              & 162                              & 161                                & 1,505                              \\ \bottomrule
\end{tabular}%
\caption{Dataset split statistics by source, detailing the distribution of chunks and Q\&A pairs.}
\label{tab:dataset_split}
\end{table}

\begin{table}[h!]
\centering
\begin{tabular}{lrrr}
\toprule
                                             & \multicolumn{1}{c}{\textbf{Min}} & \multicolumn{1}{c}{\textbf{Mean}} & \multicolumn{1}{c}{\textbf{Max}} \\ \midrule
\multicolumn{1}{l|}{Question}       & 9                                & 35                                & 82                               \\
\multicolumn{1}{l|}{Answer}         & 6                                & 56                                & 271                              \\
\multicolumn{1}{l|}{Chunk (\texttt{Ver.0})} & 66                              & 843                               & 9,528                            \\
\multicolumn{1}{l|}{Chunk (\texttt{Ver.1})} & 106                              & 884                               & 9,528                            \\ \bottomrule
\end{tabular}%
\caption{Token length statistics in the training data, processed with the BGE M3 tokenizer. \texttt{Ver.0} refers to the raw chunk without the pre-pended header, while \texttt{Ver.1} includes header.}
\label{tab:data_token_stat}
\end{table}

\begin{figure}[h!]
    \centering
    \setlength{\fboxrule}{0.2pt}
    \fbox{\includegraphics[width=1.0\linewidth]{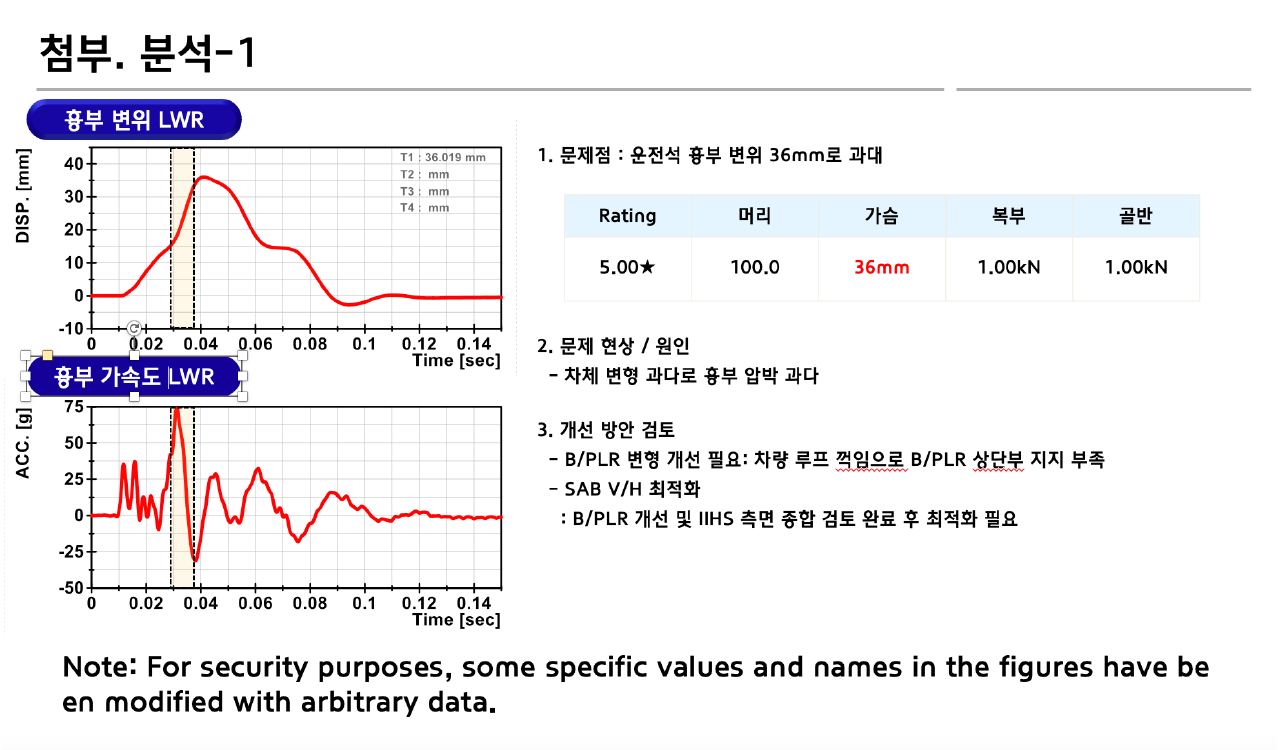}}
    \caption{Screenshot of an internal slide from an automotive crash-collision safety report, containing figures, tables, and engineering annotations for vehicle component safety.}
    \label{fig:original_slide}
\end{figure}

The original slide (Figure \ref{fig:original_slide}) often contains tables, graphs, and images, which are simplified into plain text descriptions during data processing. To evaluate performance on questions referencing these elements, we sample 143 multi-modal chunks and generated 1,861 targeted questions. Evaluation results are provided in Section \ref{sec:retriever_ranker_result_table_specific}. Additionally, we extract and prepend headers summarizing each report, including the \textit{Test Name}, \textit{Region}, \textit{State}, and \textit{Purpose}, which significantly improve the retrieval phase (See Section \ref{sec:retriever_ranker_result}). Domain experts are responsible for the entire process, including prompt engineering and reviewing intermediate and final outputs. Detailed prompt templates for QA generation, Markdown conversion, and evaluation are available in our released repository (see Section \ref{sec:conclusion}).

\subsection{RAG Components}
\label{sec:approach_components}
We strategically select backbone models and training strategies for the Retriever \& Re-ranker (Section~\ref{sec:approach_ranking}) and Generation Model (Section~\ref{sec:apporach_generation}), with a focus on effective integration under the constraints of on-premise deployment, domain adaptation, and Korean language support.

\subsubsection{Retrieval \& Re-ranking}
\label{sec:approach_ranking}
We employ a Dual-Encoder as our retriever, fine-tuning it on our training data. For a question $q$, the model retrieves the top $n$ chunks from $m$ chunks $\mathbf{D}$, based on the [CLS] token embedding similarity between $q$ and each chunk $d_i \in \mathbf{D}$. We use BGE-M3 \cite{chen-etal-2024-m3} as the backbone—a multilingual encoder capable of processing Korean—and adopt publicly available fine-tuning code~\footnote{\url{https://github.com/FlagOpen/FlagEmbedding}}. The prcocess can be described as follows:
\[
\begin{split}
q_{\text{[CLS]}} \in \mathbb{R}^{1 \times d}; \; \mathbf{D}_{\text{[CLS]}} \in \mathbb{R}^{m \times d}; \\
\text{Similarity}(q, \mathbf{D}) = q_{\text{[CLS]}} \cdot \mathbf{D}_{\text{[CLS]}}^\top \in \mathbb{R}^{1 \times m} \\
\mathbf{D}_{\text{top}_n} = \text{Sort}\left(\mathbf{D}, \; \text{key}=\text{Similarity}(q, \mathbf{D})\right)[:n]
\end{split}
\]
Our re-ranker is a point-wise Cross-encoder initialized with the weights of the fine-tuned retriever. It is trained on a classification task \cite{DBLP:journals/corr/abs-1910-14424} that takes a single $(q, d_i)$ pair as input and outputs a scalar relevance score. It re-ranks $D_{top_n}$, obtained from the retrieval stage, to extract $D_{top_k}$, which will be passed to the generation model. The process is formalized as follows:
\[
\begin{split}
\mathbf{D}_{\text{top}_n} = [d_1, d_2, \dots, d_n] \\
x_i = \text{"\{$q$\} \{sep\_token\} \{$d_i$\}"} \quad \forall d_i \in \mathbf{D}_{\text{top}_n} \\
\text{Rel}_{x_i} = \text{Ranker}(x_i) \quad \forall i \in \{1, 2, \dots, n\} \\
\mathbf{D}_{\text{top}_k} = \text{Sort}(\mathbf{D}_{\text{top}_n}, \; \text{key}=\text{Rel}_{x_i})[:k]
\end{split}
\]
In domain-specific settings where the relevance of specific entities (e.g., car models, test types, or component names) must be emphasized, re-ranking should account for not only general semantic similarity between queries and documents but also fine-grained, term-level relevance. To optimize $\text{Rel}_{x_i}$, we apply Token Selection (TS) + Term Control Layer (TCL) training method from RRADistill~\cite{choi2024rradistilldistillingllmspassage}. It effectively integrates the importance of the general semantic and specific query terms during the training process.
\[
\begin{split}
\begin{aligned}
d_i^{\text{TS}} = \text{RRA}_\text{TS}(q, d_i) \\
x_i^{\text{TS}} = \text{"\{$q$\} \{$sep\_token$\} \{$d_i^{\text{TS}}$\}"} \\
\text{Rel}_{x_i}^{\text{TS}} = \text{Ranker}(x_i^{\text{TS}}) \quad \forall i \in \{1, 2, \dots, n\} \\
\mathcal{L} = \frac{1}{n} \sum_{i=1}^{n} \left( \alpha \cdot \mathcal{L}(\text{Rel}_{x_i}, y_i) + \beta \cdot \mathcal{L}(\text{Rel}_{x_i}^{\text{TS}}, y_i) \right)
\end{aligned}
\end{split}
\]
TS+TCL are used during training only, while inference uses the standard Cross-Encoder only with $\text{Rel}_{x_i}$. The label $y$ is binary: $y = 1$ for relevant pairs and $y = 0$ for irrelevant pairs. Negative sampling retrieves the top 10 chunks from the train data for each query $q$, excluding the gold chunk, and randomly selects three from the rest.

\subsubsection{Answer Generation} 
\label{sec:apporach_generation}
Our answer generation model takes the user question $q$ and the top $k$ chunks ($\mathbf{D}_{\text{top}_n}$) from the retrieval and ranking phase as input to generate the final answer $a'$, as follow:
\[
\begin{split}
a' = \text{LLM}(q, \mathbf{D}_{\text{top}_n})
\end{split}
\]
We fine-tune open-source LLMs on our training dataset, which consists of $(q, a)$ pairs derived from our data generation pipeline (Section \ref{sec:data_processing}). During fine-tuning, the question $q$ and the answer $a$ are concatenated into a single sequence $S = [q; a]$, and the model is trained in an auto-regressive manner to predict the next token. Specifically, we choose Qwen-2.5~\footnote{We fine-tune Qwen 2.5 (14B, 72B) on our training dataset, performing full fine-tuning on the 14B model and applying Low-Rank Adaptation (LoRA) \citep{hu2021loralowrankadaptationlarge} for the 72B model.} \cite{qwen} as our backbone LLMs, one of the few multilingual models that officially supports Korean. Notably, there are currently no Korean-centric open-source LLMs with sufficient context length to address our requirements.

\subsection{Reference Matching Algorithm}
\label{sec:reference_matching_algorithm}
We propose a reference matching algorithm that enables traceability to the original source documents. The algorithm segments the generated answer and links each segment to relevant reference documents using a Dynamic Programming (DP)~\cite{bellman1954dynamic} approach, leveraging our re-ranker (Section~\ref{sec:approach_ranking}). \textbf{Algorithm~\ref{algo:reference_matching}} outlines the detailed procedure. In Step 1, the algorithm computes the scores for all possible sentence subsequences $a'_{\text{segments}}$ against the top-k chunks $D_{\text{top}_k}$, using re-ranker. Here, $a'_{\text{segments}}$ refers to candidate subsequences of sentences within $a'$ considered for matching. In Step 2, it selects the optimal segment-chunk combinations, updating scores and recording the best choices for each ending sentence (e.g., if the answer consists of five sentences like $a = [s_1, \dots, s_5]$, the choices might be $(s_1\text{:}s_1, d_1), (s_2\text{:}s_2, d_3), (s_2\text{:}s_3, d_3), (s_1\text{:}s_4, d_1),$ $(s_5\text{:}s_5, d_2)$). Finally, in Step 3, backtracking is performed to retrieve and output the best matches from the choices (e.g., $(s_1\text{:}s_4, d_1), (s_5\text{:}s_5, d_2)$), i.e., the first to fourth sentences of $a'$ are grouped as one segment aligned to $d_1$, and the fifth sentence is aligned to $d_2$.

\subsection{Deployment}
\label{sec:deploy}
We internally deployed the on-premise RAG system—comprising a 0.5B retriever, a 0.5B re-ranker, and a 72B LLM— for pilot testing. Employees evaluated the system and provided positive feedback. A full-scale deployment is currently being planned.

\SetKwComment{Comment}{$\triangleright$ }{}
\SetKwInOut{Input}{Input}
\SetKwInOut{Output}{Output}
\DontPrintSemicolon
\RestyleAlgo{ruled}
\SetAlFnt{\small}
\vspace{-0.1cm}
\begin{algorithm}[t]
\caption{$\;$\textbf{Reference Matching Algorithm}}
\label{algo:reference_matching}
\Input{$a' = [s_1, \dots, s_n]$, $\mathbf{D}_{\text{top}_k} = [d_1, \dots, d_k]$}
\Output{$\{(s_i\!:\!s_j,\, d_t,\, \text{Score}_{(i,j,d_t)})\}_{i,j,t}$}
\vspace{0.1cm}

\Comment{Step 1: Compute Segment-Chunks Scores}
$a'_{\text{segments}} \gets \{ \text{concat}(s_i, \dots, s_j) \mid 1 \leq i \leq j \leq n \}$ \\
\ForEach{segment $\mathbf{seg}_u \in a'_{\text{segments}}$}{
  \ForEach{chunk $d_t \in \mathbf{D}_{\text{top}_k}$}{
    $\text{Score}_{(i,j,d_t)} \gets \text{Re-ranker}(\mathbf{seg}_u, d_t)$
  }
}

\vspace{0.1cm}
\Comment{Step 2: Optimal Score Selection}
Initialize $dp$ and $choices$ arrays \\
\ForEach{ending sentence $s_j$}{
  \ForEach{starting sentence $s_i$ with $1 \leq i < j$}{
    \ForEach{chunk $d_t \in \mathbf{D}_{\text{top}_k}$}{
      \If{$\text{Score}_{(i,j,d_t)} > dp[j+1]$}{
        $dp[j+1] \gets \text{Score}_{(i,j,d_t)}$ \\
        Record choice $(i,j,d_t)$ in $choices$
      }
    }
  }
}

\vspace{0.1cm}
\Comment{Step 3: Backtracking}
$current \gets n$
\While{$current > 0$}{
  Retrieve best $(i,j,d_t)$ from $choices$ for $current$ \\
  Add $(s_i\!:\!s_j, d_t, \text{Score}_{(i,j,d_t)})$ to result \\
  Set $current \gets i$ to move to previous segment.
}
\Return{result}
\end{algorithm}

\section{Experiment}
\subsection{Retrieval \& Re-ranking}
\label{sec:retrieval_ranking}
\subsubsection{Evaluation}
We evaluate the retriever and re-ranker using Mean Average Precision (MAP@k) and Success@k \cite{Manning2008IR}. MAP@k measures the ranking quality by averaging precision over results up to rank $k$, while Success@k indicates the proportion of queries with at least one relevant result in the top $k$. The high Success@k score indicates that relevant chunks are retrieved, while the improved MAP@k highlights better ranking of those retrieved chunks. The evaluation is conducted using test set questions, but the chunk pool for retrieval included all splits (training, validation, and test) to ensure sufficient data and avoid biases from the limited test set size.

\subsubsection{Result}
\label{sec:retriever_ranker_result}
Table \ref{tab:retriever_performance} shows that prepending a header to each chunk—adding only a small number of tokens (see Table \ref{tab:data_token_stat})—significantly enhances retrieval performance. Fine-tuning the BGE-M3 model on \texttt{Ver.1} (BGE-FT) further improves results notably, demonstrating the importance of task-specific model adaptation to optimize performance. Table \ref{tab:ranker_performance} shows that the re-ranking phase notably enhances the precision of retrieved information. The results indicate that ranking performance is optimal when the number of retrieved chunks is limited to 10. Despite the constraints of retriever failure, this improves retrieval results by about $+$4\% in terms of MAP@1. In addition, our ablation study confirmed that applying TS+TCL during training slightly improves ranking performance.

\begin{table}[h!]
\centering
\begin{tabular}{cllrrr}
\toprule
\textbf{Model}                                                                              & \multicolumn{1}{c}{\textbf{Train}}           & \multicolumn{1}{c}{\textbf{Test}} & \multicolumn{1}{c}{\textbf{MAP@1}} & \multicolumn{1}{c}{\textbf{MAP@5}} & \multicolumn{1}{c}{\textbf{MAP@10}} \\ \midrule
\multicolumn{1}{c|}{\multirow{2}{*}{\begin{tabular}[c]{@{}c@{}}BGE-\\ Vanilla\end{tabular}}} & \multicolumn{1}{l|}{\multirow{2}{*}{N/A}}    & \multicolumn{1}{l|}{\texttt{Ver.0}}       & 0.1978                             & 0.2672                             & 0.2766                              \\
\multicolumn{1}{c|}{}                                                                       & \multicolumn{1}{l|}{}                        & \multicolumn{1}{l|}{\texttt{Ver.1}}       & 0.2991                             & 0.3978                             & 0.4091                              \\
\midrule
\multicolumn{1}{c|}{\multirow{1}{*}{BGE-FT}} & \multicolumn{1}{l|}{\multirow{1}{*}{\texttt{Ver.1}}} & \multicolumn{1}{l|}{\texttt{Ver.1}}       & \textbf{0.6048}                             & \textbf{0.7111}                             & \textbf{0.7175}                              \\ \bottomrule
\end{tabular}%
\caption{Retrieval performance across models with different data configurations. \texttt{Ver.0} refers to the raw chunk without a prepended header, while \texttt{Ver.1} includes the header.}
\label{tab:retriever_performance}
\vspace{-1.2em}
\end{table}

\begin{table}[h!]
\centering
\centering
\begin{tabular}{c|cccc} 
\toprule
\textbf{$k$} & \textbf{MAP@1}                              & \textbf{MAP@5}                              & \textbf{MAP@10}                             & \textbf{Success@5}                           \\ \midrule
10 & \textbf{0.6444} \scriptsize(± 0.01) & \textbf{0.7416} \scriptsize(± 0.01) & \textbf{0.7464}  \scriptsize(± 0.01) & \textbf{0.8839} \scriptsize(± 0.00) \\
20 & 0.6362 \scriptsize(± 0.01) & 0.7353 \scriptsize(± 0.01) & 0.7413 \scriptsize(± 0.01) & 0.8822 \scriptsize(± 0.01) \\
30 & 0.6328 \scriptsize(± 0.02) & 0.7316 \scriptsize(± 0.02) & 0.7378 \scriptsize(± 0.02) & 0.8796 \scriptsize(± 0.01) \\ 
\bottomrule
\end{tabular}%
\caption{Ranking performance based on the number of retrieved chunks $k$. Instances that failed during the retrieval phase were scored as 0.}
\vspace{-1.2em}
\label{tab:ranker_performance}
\end{table}

\subsubsection{Multi-modal Specific Questions}
\label{sec:retriever_ranker_result_table_specific}
Table \ref{tab:ranker_performance_mutimodal_specific} summarizes the performance on questions requiring information from tables, graphs, and images. Compared to general questions (Section \ref{sec:retriever_ranker_result}), the results are notably lower, with the retriever facing significant challenges. Even at $k$ = 30, Success@k only reaches 83.61\%, underscoring a major bottleneck in the retrieval phase. Despite these challenges, the re-ranker proved effective, delivering a substantial performance improvement of approximately $+$15\% in MAP@1 at $k$ = 20. These results underscore the need for further advancements to better handle multi-modal questions.

\begin{table}[h!]
\centering
\centering
\begin{tabular}{l|c|rrr}
\toprule
\multicolumn{1}{c|}{\textbf{Phase}}                                    & \textbf{$k$} & \multicolumn{1}{c}{\textbf{MAP@1}} & \multicolumn{1}{c}{\textbf{MAP@5}} & \multicolumn{1}{c}{\textbf{MAP@10}} \\ \midrule
\begin{tabular}[c]{@{}l@{}}\textbf{Retrieval}\\ \scriptsize{w/ BGE-FT}\end{tabular} & N/A        & 0.3482                             & 0.4561                             & 0.4684                              \\ \midrule
\multirow{3}{*}{\textbf{Reranking}}                                    & 10         & 0.4723                             & 0.5587                             & 0.5630                              \\
                                                                       & 20         & \textbf{0.4949}                    & 0.5923                             & 0.5995                              \\
                                                                       & 30         & 0.4933                             & \textbf{0.5965}                    & \textbf{0.6041}                     \\ \bottomrule
\end{tabular}%
\caption{Performance of retrieval and ranking on multi-modal specific questions. The same models and evaluation methods described in Section \ref{sec:retriever_ranker_result} were used.}
\label{tab:ranker_performance_mutimodal_specific}
\vspace{-1em}
\end{table}

\subsubsection{Analysis}
\label{subsec:retriever_ranker_analysis}
\begin{figure}[h!]
    \centering
    \includegraphics[width=0.9\linewidth]{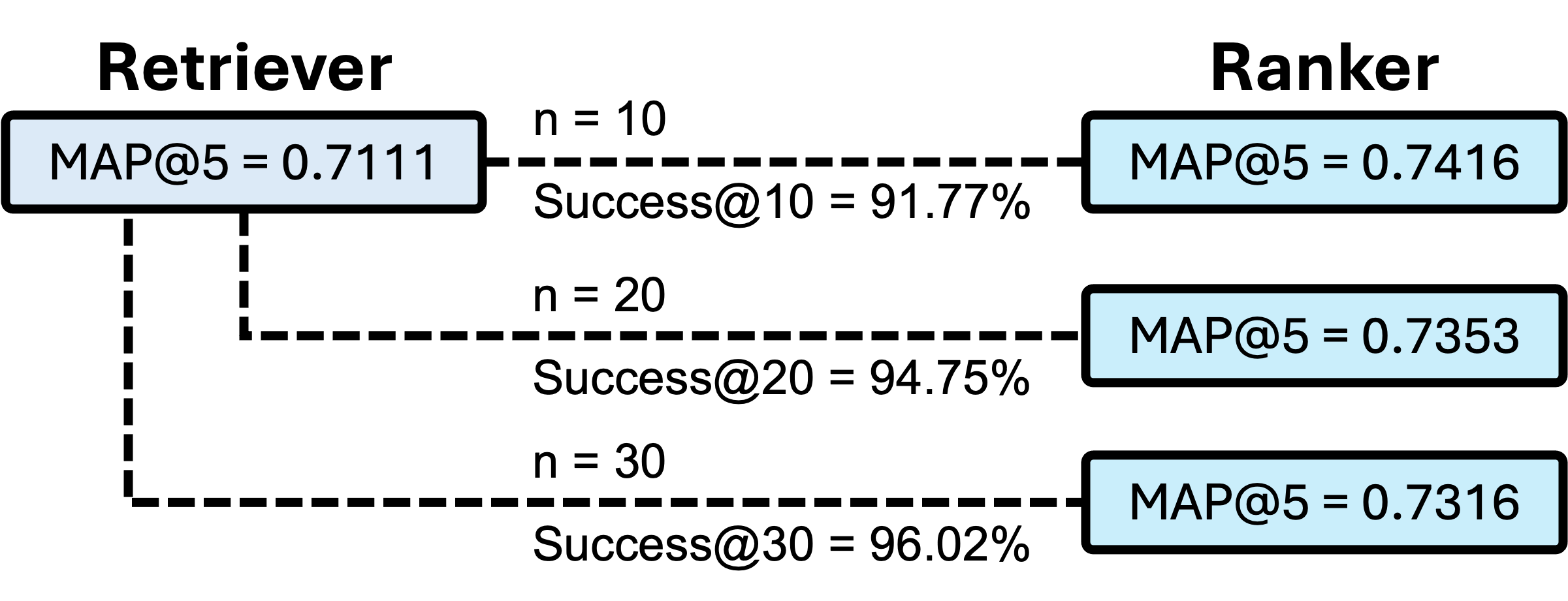}
    \caption{The impact of the number of retrieved chunks $n$ on retrieval success rates and ranking performance.}
    \label{fig:retriever_ranker}
\vspace{-1em}
\end{figure}

\begin{figure}[h!]
    \centering
    \includegraphics[width=0.9\linewidth]{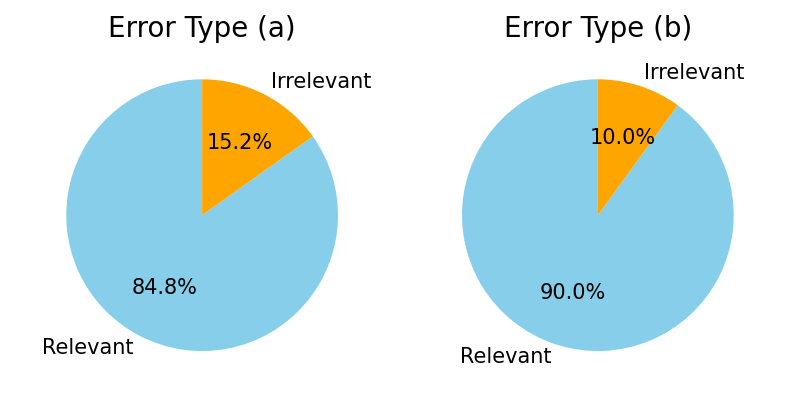}
    \caption{Human evaluation of the re-ranker top-1 for each error type, based on 100 sampled cases, assessing whether it is relevant to the given question.}
    \label{fig:retrieval_ranking_error_type}
\vspace{-1em}
\end{figure}

Figure \ref{fig:retriever_ranker} illustrates the trade-off between retrieval success and re-ranking performance as the number of retrieved chunks $n$ increases. While increasing $n$ improves the retrieval success rate (e.g., from 92\% at $n$ = 10 to 96\% at $n$ = 30), it can lead to diminishing improvements in re-ranking performance (e.g., MAP@5 decreases from 0.74 at $n$ = 10 to 0.73 at $n$ = 30). This result highlights the importance of carefully selecting $n$ to balance retrieval success and re-ranking quality, as overly increasing $n$ may not yield proportional benefits for re-ranking. In addition, an analysis of the failed cases during the retrieval and ranking phase identified two error types: (a) Top-10 retrieval failure (23.3\%) and (b) Retrieval success but incorrect re-ranker top-1 (76.7\%). Type (a) were mainly caused by the open-ended nature of the questions, which led to multiple relevant documents beyond the gold context. In contrast, type (b) occurred with more specific and detailed questions, where there were several relevant chunks from the same vehicle crash collision test, in addition to the gold chunk, making it difficult to identify a single correct one. Failures in both types were often due to the presence of multiple valid chunks, rather than to the ranking of irrelevant chunks. Table \ref{fig:retrieval_ranking_error_type} presents a human evaluation of 100 sampled cases for each error type, assessing whether the re-ranker top-1 was relevant to the given question, even though it did not match the gold chunk.

\subsection{Answer Generation}
\label{sec:answer_generation}

\subsubsection{Evaluation}
\label{subsec:rag_evaluation}
We employ the \textbf{LLM-as-a-judge} approach \citep{zheng2023judgingllmasajudgemtbenchchatbot} to evaluate the performance of both LLM-only and RAG models. For the LLM evaluation, we compare four model variations: the vanilla and fine-tuned versions of 14B and 72B models. GPT-4o~\footnote{\url{https://platform.openai.com/docs/models/gpt-4o}} is used to rank anonymized models (A, B, C, and D) based on factual correctness, helpfulness, and informativeness. These metrics are chosen to evaluate distinct, non-overlapping dimensions:
\begin{itemize}
    \item \textbf{Correctness} measures alignment with the gold answer, rewarding correct responses without hallucinations.
    \item \textbf{Helpfulness} assesses clarity and relevance in addressing key points, while avoiding unnecessary content.
    \item \textbf{Informativeness} evaluates the inclusion of relevant details or additional context that enhances completeness.
\end{itemize}

\noindent For RAG evaluation, we use the best-performing one among four models and compare its performance with and without RAG integration. Instead of pairwise comparisons, we employ single-answer evaluations, scoring responses (1–5) on three aspects. This approach offers a more detailed assessment, emphasizing how RAG impacts specific aspects of response quality. To assess the reliability of LLM evaluation, we conducted a \textbf{human evaluation} by a domain expert in automotive safety. inter-annotator agreement (IAA) could not be calculated due to resource constraints, and the evaluation was limited to 100 randomly selected and anonymized samples; scaling to more annotators and larger sets will be part of future work.

\subsubsection{Result}
\begin{figure}[h!]
    \centering
    \includegraphics[width=1.0\linewidth]{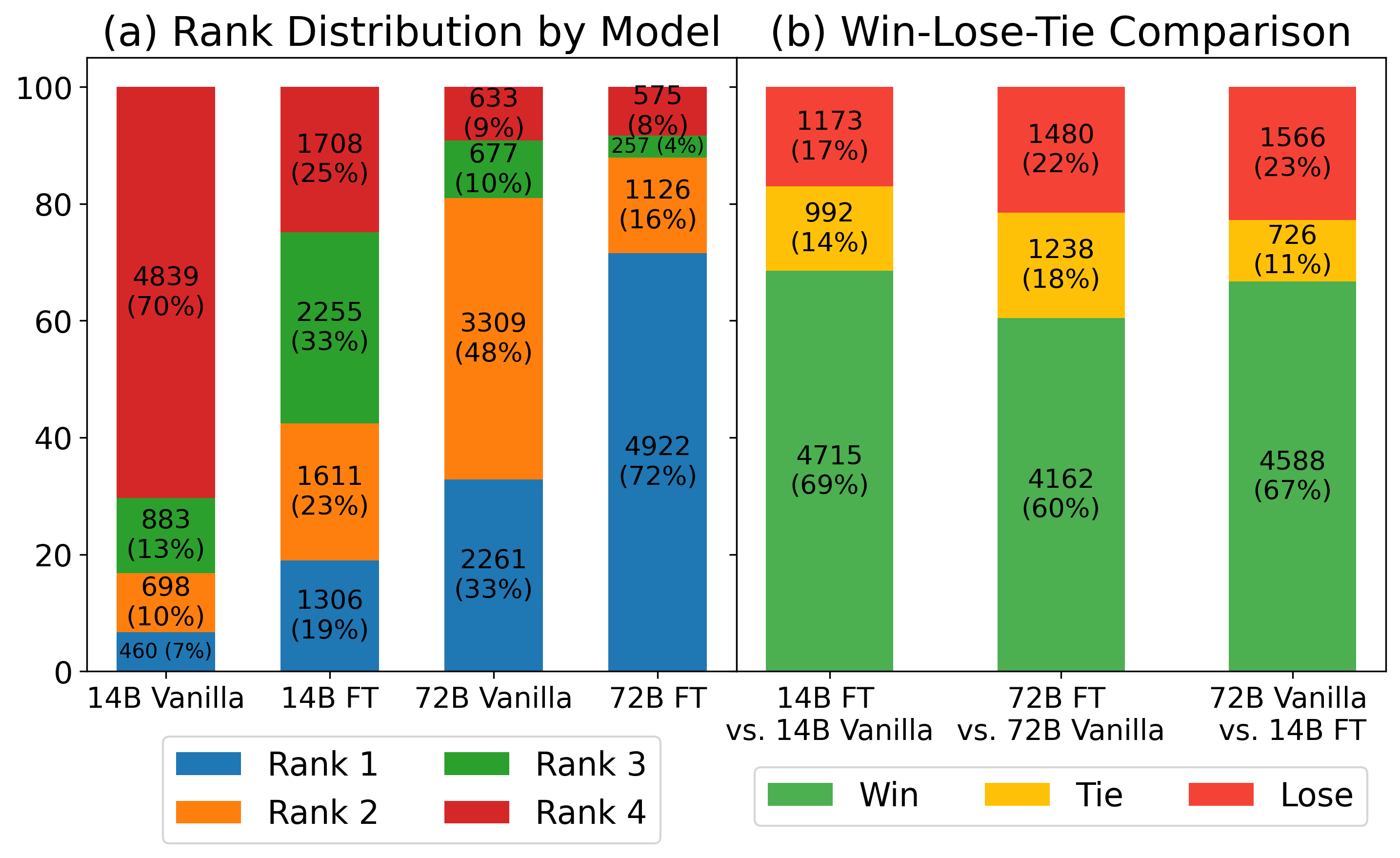}
    \caption{LLM-only performance comparison: \textbf{(a)} Proportion of responses ranked 1st–4th across the four models; \textbf{(b)} Win, tie, and loss rates for model pairs, where a "win" means the first model outperformed the other.}
    \label{fig:4models_rank}
\vspace{-0.5em}
\end{figure}

Figure \ref{fig:4models_rank} illustrates the rank distribution and win-lose-tie comparison across four Qwen models. Among them, the 72B fine-tuned one shows the best performance, followed by the 72B vanilla, 14B fine-tuned, and 14B vanilla models. These results emphasize the critical impact of model size and demonstrate the significant effectiveness of fine-tuning, as fine-tuned models consistently outperform their vanilla counterparts.

\begin{figure}[h!]
    \centering
    \includegraphics[width=0.9\linewidth]{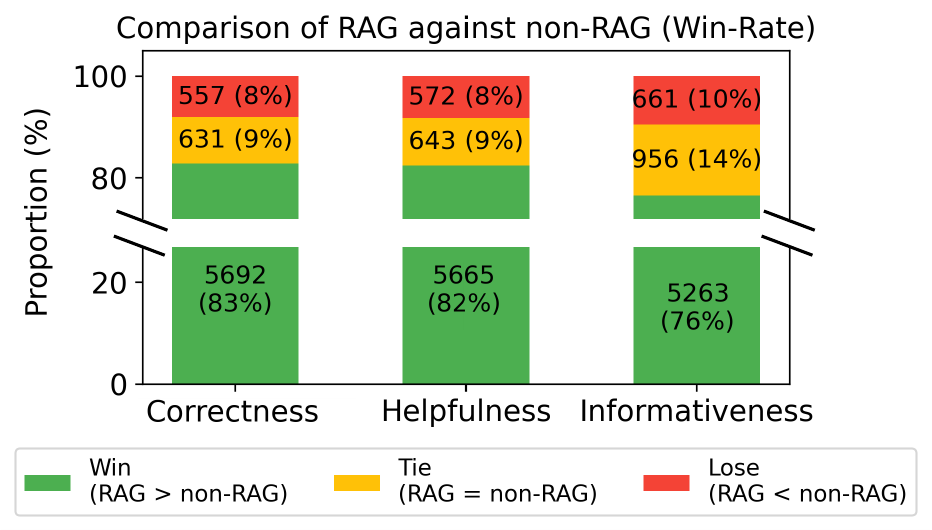}
    \caption{Win-Tie-Lose comparison of the model with and without RAG across three evaluation metrics.}
    \label{fig:rag_ft_model_winrate}
\vspace{-0.5em}
\end{figure}

\noindent We compare the performance of the 72B fine-tuned model in its LLM-only (non-RAG) and RAG-integrated versions. Table \ref{tab:rag-nonrag-avg} compares the average scores for correctness, helpfulness, and informativeness between the two versions, while Figure \ref{fig:rag_ft_model_winrate} shows the win rates. The results clearly demonstrate that integrating RAG with the LLM improves factual correctness by reducing hallucinations and provides more helpful and informative answers.

\begin{table}[h!]
\centering
\begin{tabular}{lrr}
\toprule
\textbf{Metric} & \textbf{non-RAG} & \textbf{RAG} \\
\midrule
\textbf{Correctness} & 2.18 & \textbf{4.12} \textcolor{blue}{(+1.94)} \\
\textbf{Helpfulness} & 2.52 & \textbf{4.19} \textcolor{blue}{(+1.67)} \\
\textbf{Informativeness} & 2.61 & \textbf{3.77} \textcolor{blue}{(+1.16)} \\
\bottomrule
\end{tabular}%
\caption{Average scores (1 to 5 scale) for the model with and without RAG across three metrics.}
\label{tab:rag-nonrag-avg}
\vspace{-1em}
\end{table}

\begin{figure}[h!]
    \centering
    \includegraphics[width=1.0\linewidth]{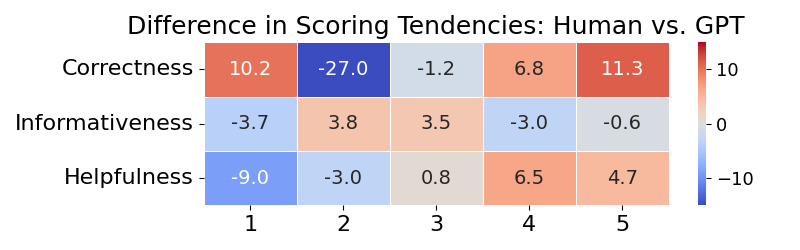}
    \caption{Heatmap showing differences in score distributions between human and GPT evaluations, computed as the percentage of scores (1–5) given by humans minus those given by GPT. \textcolor{orange}{Positive values} mark scores more frequent in human, and \textcolor{blue}{negative values} mark scores more frequent in GPT.}
    \label{fig:heatmap}
\vspace{-1em}
\end{figure}

\noindent Table \ref{tab:human_eval_rag_vs_no_rag} shows that human evaluation aligns with GPT evaluation, both assigning higher scores to the RAG model across all aspects. Figure \ref{fig:heatmap} visualizes the differences in scoring tendencies between human and GPT evaluations. Notably, correctness and informativeness show an inverted pattern: human favors more polarized correctness scores (1 and 5), while GPT prefers mid-range (2 and 3). Additionally, human assigns higher helpfulness scores more frequently, as seen in the positive values for 4 and 5. While differences in score distributions exist between evaluators, these differences are independent of the specific models. Both evaluations consistently show that RAG outperforms non-RAG by achieving higher scores.

\begin{table}[h!]
\centering
\begin{tabular}{l|c|r|rrrrr}
\toprule
\textbf{Metric} & \textbf{Model} & \textbf{\begin{tabular}[c]{@{}c@{}}Avg. \\ Score\end{tabular}} & \textbf{1} & \textbf{2} & \textbf{3} & \textbf{4} & \textbf{5}  \\
\midrule
\multirow{2}{*}{Correctness} & A & 2.54 & 34 & 19 & 20 & 13 & 14  \\
                          & B & \textbf{4.33} & 10 & 4  & 4  & 7  & 75  \\
\midrule
\multirow{2}{*}{Helpfulness} & A & 2.89 & 2  & 38 & 36 & 17 & 7  \\
                             & B & \textbf{4.22} & 3  & 9  & 6  & 27 & 55  \\
\midrule
\multirow{2}{*}{Informativeness} & A & 2.60 & 3  & 44 & 44 & 8  & 1   \\
                                 & B & \textbf{3.68} & 2  & 8  & 31 & 38 & 21  \\
\bottomrule
\end{tabular}%
\caption{Human evaluation result comparing non-RAG (anonymized as A) and RAG (anonymized as B).}
\label{tab:human_eval_rag_vs_no_rag}
\vspace{-1em}
\end{table}

\subsection{Reference Matching}
\label{sec:reference_matching}
\begin{figure*}[t!]
\centering
\renewcommand{\arraystretch}{1.15}
\resizebox{\textwidth}{!}{%
\begin{tabular}{cclc}
\toprule
\multicolumn{4}{c}{\textbf{Question:} Has there been an example where improving the vehicle structure prevented the fracture of components?} \\ \midrule
\multicolumn{1}{c|}{\textbf{Segment 1}} & \textbf{Sentence 1} & \multicolumn{1}{l|}{Yes, there have been cases where improving the vehicle structure prevented the fracture of components.} & \textbf{Chunk 1,2,3} \\ \midrule
\multicolumn{1}{c|}{\multirow{2}{*}{\textbf{Segment 2}}} & \textbf{Sentence 2} & \multicolumn{1}{l|}{First, in the 'HDC EV' project, the design of the front bumper beam (FRT B/Beam) was changed to prevent fracture.} & \multirow{2}{*}{\textbf{Chunk 2}} \\ \cline{2-3}
\multicolumn{1}{c|}{} & \textbf{Sentence 3} & \multicolumn{1}{l|}{\begin{tabular}[c]{@{}l@{}}In the T-CAR model, fracture occurred in the FRT B/Beam, but in the Proto model, it was modified to a closed \\ cross-section shape, preventing fracture.\end{tabular}} & \\ \midrule
\multicolumn{1}{c|}{\multirow{3}{*}{\textbf{Segment 3}}} & \textbf{Sentence 4} & \multicolumn{1}{l|}{\begin{tabular}[c]{@{}l@{}}Additionally, in the 'HGC' project, body structure improvements were made to reduce the risk of deformation \\ and fracture of several components.\end{tabular}} & \multirow{3}{*}{\textbf{Chunk 1}} \\ \cline{2-3}
\multicolumn{1}{c|}{} & \textbf{Sentence 5} & \multicolumn{1}{l|}{For example, reinforcing the upper part of the A-pillar and the COWL BAR improved bending deformation.} & \\ \cline{2-3}
\multicolumn{1}{c|}{} & \textbf{Sentence 6} & \multicolumn{1}{l|}{Adding UPR HINGE B/H beads and other deformation-guiding methods reduced opening misalignment.} & \\ \midrule
\multicolumn{1}{c|}{\textbf{Segment 4}} & \textbf{Sentence 7} & \multicolumn{1}{l|}{\begin{tabular}[c]{@{}l@{}}Finally, in the 'QYZ' project, a plan was proposed to improve the tendency of fracture by adding reinforcements \\ to the S/MBR RR lower end.\end{tabular}} & \textbf{Chunk 3} \\ \midrule
\multicolumn{1}{c|}{\textbf{Segment 5}} & \textbf{Sentence 8} & \multicolumn{1}{l|}{All of these improvements aim to optimize the body structure and prevent component fracture during collisions.} & \textbf{Chunk 1,2,3} \\ \bottomrule
\end{tabular}%
}
\caption{An example where multiple chunks need to be referred to in order to answer the question.}
\vspace{-0.5em}
\label{fig:reference_case_2}
\end{figure*}

\subsubsection{Evaluation}
To evaluate the reference matching task, we randomly sample 100 triplets ($q, \mathbf{D}_{\text{top}_n}, a'$) from the test set and manually annotate to identify which chunk ($d_i \in \mathbf{D}_{\text{top}_n}$) each sentence in the generated answer ($a'$) referenced. Each sentence is annotated with one or more chunks, allowing for multiple references. We use sentence-level precision as the evaluation metric, defined as the proportion of sentences where the model prediction is included in the annotated reference set for an answer, averaged over the entire evaluation set. Since references are identified at the level of concatenated sentences, selecting the best match for the grouped information becomes most important. Evaluating recall would unfairly penalize valid predictions that do not select every reference.

\subsubsection{Result}
\label{subsec:reference_matching_result}
\begin{figure}[h!]
    \centering
    \includegraphics[width=0.9\linewidth]{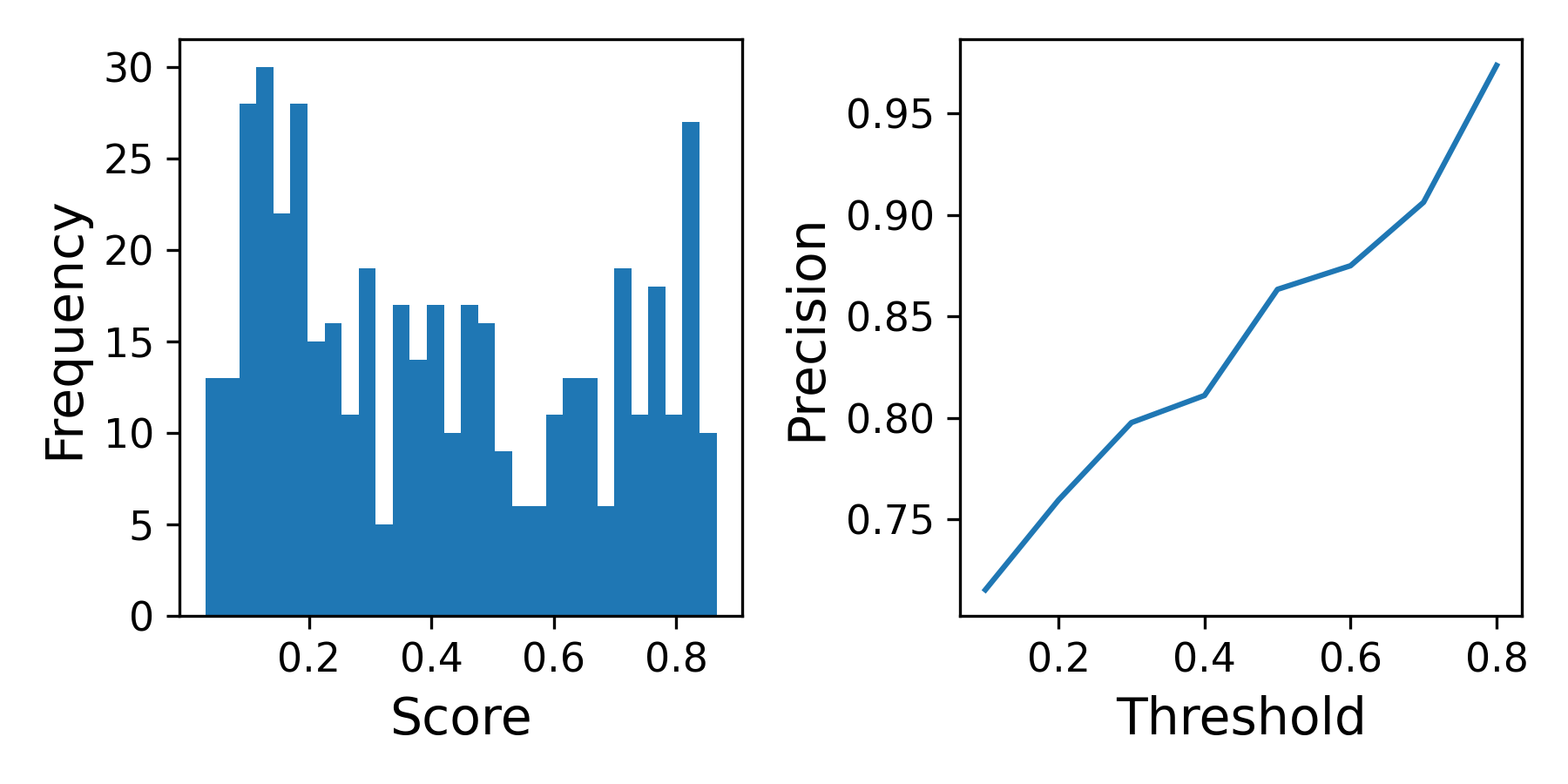}
    \caption{Distribution of matching scores from our reference matching algorithm (left) and the relationship between precision and the score threshold (right). Increasing the threshold leads to a proportional improvement in matching precision.}
    \label{fig:ref_threshold}
\vspace{-0.5cm}
\end{figure}

Table \ref{tab:result_ref} compares our reference matching (Section~\ref{sec:reference_matching_algorithm}) performance without thresholding and the fine-tuned Qwen 72B~\footnote{Prompt used for Qwen 72B reference matching is like: \textit{You are an expert tasked with analyzing user questions, an answer split into sentences, and a list of $n$ related chunks. Your job is to identify the single most relevant chunk for each sentence based on sentence indices provided. Map the segments to relevant chunks.}}, while Figure \ref{fig:ref_threshold} illustrates the score distribution and the increase in precision with score thresholding for our algorithm. Without thresholding, our algorithm achieves a sentence-level precision of 0.72, while the LLM reaches 0.81, showing a noticeable gap in performance. However, when thresholding is applied, the precision reaches 0.86 at a threshold of 0.5, with performance improving proportionally as the threshold is increased further. This demonstrates the reliability of our algorithm, allowing for controlled adjustment of reference matching quality through threshold selection. Our re-ranker used in the matching algorithm has only 0.5B parameters, compared to the LLM's 72B, demonstrating that our algorithm can achieve impressive performance with a smaller model and faster inference speed. The answers are divided into 2-3 segments on average, and we identify two types of reference matching: one where all the necessary information to answer the question is contained within a single chunk, resulting in a single-segment answer, and another where information from multiple chunks is needed to answer the question, leading to a multi-segment answer. The first type often occurs in factual questions related to specific vehicle crash collision test, while the second type is more common in open-ended questions that require referencing multiple tests (Figure \ref{fig:reference_case_2}).

\begin{table}[h!]
\centering
\begin{tabular}{c|rrr}
\toprule
\textbf{Method}                                              & \multicolumn{1}{c}{\textbf{Segment}}                       & \multicolumn{1}{c}{\textbf{\begin{tabular}[c]{@{}c@{}}Sentence-level \\ Precision\end{tabular}}} & \multicolumn{1}{c}{\textbf{\begin{tabular}[c]{@{}c@{}}Inference\\ Time\end{tabular}}} \\ \midrule
Ours & 2.7 ($\pm$ 1.5) & 0.72 ($\pm$ 0.27) & 3.92 \\ \midrule
LLM & 1.9 ($\pm$ 0.9) & 0.81 ($\pm$ 0.26) & 13.54 \\ \bottomrule
\end{tabular}%
\caption{Performance comparison of our re-ranker leveraged algorithm (without thresholding) and the fine-tuned Qwen 72B. \textbf{Segment} is the average number of segments per answer, \textbf{Sentence-level Precision} is the average proportion of successfully matched sentences, and \textbf{Inference Time} is the average time (in sec.) to complete reference matching for one answer.}
\label{tab:result_ref}
\vspace{-0.5em}
\end{table}

\section{Conclusion}
\label{sec:conclusion}
In this study, we presented an on-premise RAG-QA system tailored for internal corporate documents, with a focus on the automotive domain. The system comprises a domain-adapted retriever, re-ranker, LLM, and reference matcher, and has proven effective through comprehensive evaluations and a pilot test with internal employees. Ongoing work addresses current limitations, including the limited availability of open-source LLMs that support Korean and long-context processing. We are experimenting with newly released models such as Qwen2.5-1M \cite{yang2025qwen251mtechnicalreport}, which may serve as viable alternatives. To better handle multi-modal questions involving tables and figures, we are also exploring LLaMA 3.2 Vision \cite{grattafiori2024llama3herdmodels} to enhance the textual representation of visual content. We believe our RAG development pipeline offers a practical solution for organizations seeking to effectively utilize fragmented internal documents in decision-making processes. We released the code—including data processing and RAG components training—as well as all the prompts used: \textit{\url{https://github.com/emorynlp/MultimodalRAG}}

\begin{acks}
We gratefully acknowledge the support of Hyundai Motor Company. Any opinions, findings, conclusions, or recommendations expressed in this material are those of the authors and do not necessarily reflect the views of Hyundai Motor Company.
\end{acks}

\section*{GenAI Usage Disclosure}
We used ChatGPT~\footnote{https://chatgpt.com/} solely for minor English grammar and style corrections
after the authors had drafted the full text.


\bibliographystyle{ACM-Reference-Format}
\bibliography{sample-base}

\appendix

\end{document}